\documentclass[conference, letterpaper, 10 pt]{IEEEtran}
\IEEEoverridecommandlockouts
\usepackage[letterpaper, left=0.75in, right=0.75in, top=0.75in, bottom=0.75in]{geometry}
\usepackage{amsmath,amssymb,amsfonts}
\usepackage[english]{babel} 
\usepackage{blindtext}
\usepackage{subcaption}
\usepackage{algorithmic}
\usepackage{hyperref}
\usepackage{cleveref}
\usepackage{float}
\crefname{figure}{Fig.}{Figs.}
\usepackage{graphicx}
\usepackage{textcomp}
\usepackage{makecell}
\usepackage{booktabs}
\usepackage{xcolor}

\usepackage[acronym]{glossaries}
\makeglossaries

\usepackage{cite}

\def\BibTeX{{\rm B\kern-.05em{\sc i\kern-.025em b}\kern-.08em
    T\kern-.1667em\lower.7ex\hbox{E}\kern-.125emX}}

\usepackage{blindtext}

\usepackage{hyperref}
\usepackage{cleveref}
\crefname{figure}{Fig.}{Figs.}

\makeatletter
\def\@maketitle{%
  \newpage
  \null
  \vskip 2em
  \begin{center}%
    {\LARGE\bfseries \@title \par} 
    \vskip 1em
     {\large
       Christian Westerdahl, Jonas Poulsen, Daniel Holmelund, Peter Nicholas Hansen, Fletcher Thompson, Roberto Galeazzi \par
     }
    \vskip 1em
  \end{center}%
  \par
  \vskip 1em
}
\makeatother

\begin{document}

\title{Seabed-to-Sky Mapping of Maritime Environments with a Dual Orthogonal SONAR and LiDAR Sensor Suite
}

\maketitle

\begin{abstract}

Critical maritime infrastructure increasingly demands situational awareness both above and below the surface, yet existing ``seabed-to-sky'' mapping pipelines either rely on GNSS (vulnerable to shadowing/spoofing) or expensive bathymetric sonars. We present a unified, GNSS-independent mapping system that fuses LiDAR–IMU with a dual, orthogonally mounted Forward Looking Sonars (FLS) to generate consistent seabed-to-sky maps from an Autonomous Surface Vehicle. On the acoustic side, we extend orthogonal wide-aperture fusion to handle \emph{arbitrary inter-sonar translations} (enabling heterogeneous, non co-located models) and extract a \emph{leading edge} from each FLS to form line-scans. On the mapping side, we modify LIO-SAM to ingest both stereo-derived 3D sonar points and leading-edge line-scans \emph{at and between} keyframes via motion-interpolated poses, allowing sparse acoustic updates to contribute continuously to a single factor-graph map. We validate the system on real-world data from Belvederekanalen (Copenhagen), demonstrating real-time operation with $\sim$2.65\,Hz map updates and $\sim$2.85\,Hz odometry while producing a unified 3D model that spans air–water domains.
\end{abstract}


\section{Introduction}

The vulnerability of European critical maritime infrastructure and the threats against them from both state and non-state actors as well as natural hazards have become starkly evident. Examples include the explosions on the Nord Stream natural gas pipelines in September 2022~\cite{adomaitis_what_2025}, the damage to the BCS East-West Interlink and C-Lion1 fiber-optic cables in November 2024~\cite{kottasova_two_2024}, and most recently, the disruption of the Sweden-Latvia submarine cable in January 2025~\cite{sytas_sweden_2025}. These events highlighted how the safety and resilience of assets such as communication cables, offshore wind farms, and energy pipelines are not only vital for European economies but also for ensuring global connectivity and security. 

While current maritime domain awareness systems are based on ship routing and position given from the ships themselves, augmented by data from coastal radars, CCTV, and patrols, the main threats might come from covert underwater operations. Hence, investments in underwater sensors and drones are warranted to achieve subsea surveillance~\cite{bueger_critical_2023}.

In response, the European Union has strengthened its policy framework through the revised EU Maritime Security Strategy~ \cite{noauthor_maritime_2021} and, more concretely, the 2025 Action Plan on Cable Security~\cite{noauthor_eu_2025}, which emphasizes prevention, detection, response, and deterrence as the cornerstones of resilience for critical underwater infrastructure. Parallel to EU initiatives, NATO has established the Critical Undersea Infrastructure Coordination Cell~\cite{noauthor_nato_2023} and launched the Maritime Centre for the Security of Critical Undersea Infrastructure~\cite{noauthor_nato_2024}, underscoring that hybrid threats against seabed and port assets have become central to Euro-Atlantic defense agendas.

Within this evolving strategic landscape, ensuring consistent surveillance both above and below the water surface becomes vital. In the underwater domain, Remotely Operated Vehicles (ROVs) have long been utilized for inspection and mapping tasks. \cite{roman_improved_2005} demonstrate how the bathymetry submaps, obtained from a multibeam S Sound and Ranging (SONAR) device, were fused to provide a more accurate location and trajectory estimate for the vehicle. To extend these mapping capabilities without reliance on tethers, Autonomous Underwater Vehicles (AUVs) have been employed. Building on SONAR based mapping approaches, in \cite{ribas_underwater_2008} and \cite{loi_sonar_2024}, SONAR devices were mounted on an Autonomous Underwater Vehicles to obtain maps of man-made structured underwater environments. \cite{wang_virtual_2022} introduce the concept of a \textit{virtual map}, including uncertainty in the maps, thus identifying areas for further autonomous exploration.
\begin{figure}[tbp]
\centerline{\includegraphics[width=\columnwidth]{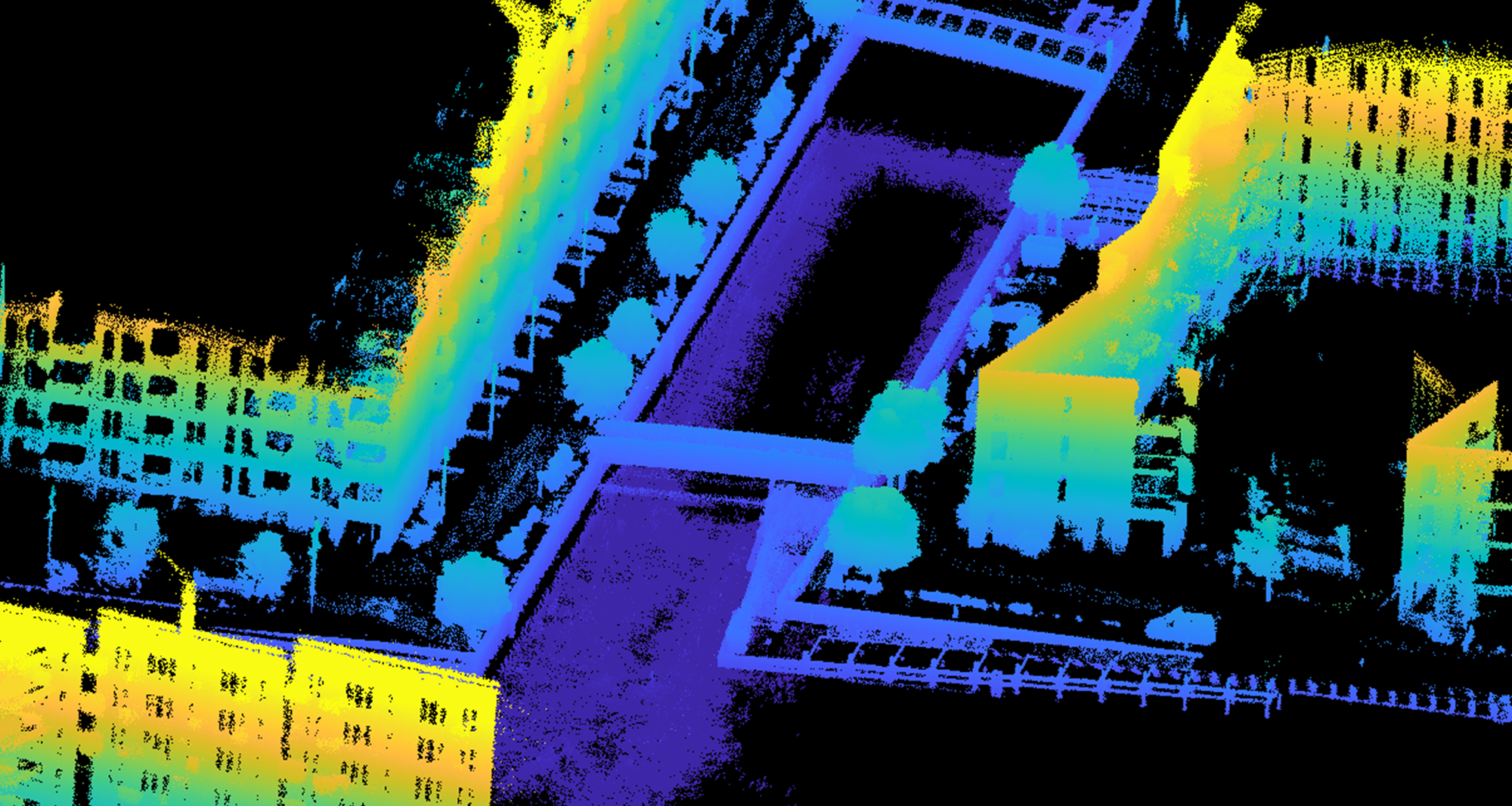}}
\caption{The obtained ``seabed-to-sky'' map from data collected in Belvederekanalen, Sydhavnen, Copenhagen on the 5th of December 2024.}
\label{fig:sonar_bed}
\vspace{-4mm} 
\end{figure}

On the surface, autonomous surface vehicles (ASVs) offer several operational advantages: they have access to the Global Navigation Satellite System (GNSS) and, equipped with multimodal sensor platforms, they can extend the mapping capabilities to encompass the environment both above and below water. This was first proposed in \cite{leedekerken_mapping_2014}, where an ASV equipped with both LiDAR and SONAR devices recorded both environments simultaneously, while utilizing the GNSS, an Inertial Measurement Unit (IMU) and a Doppler Velocity Log (DVL) for accurate positioning. In the case of GNSS denied environments, \cite{han_coastal_2019} employed a marine RADAR in a Simultaneous Localization and Mapping (SLAM) framework, in which the pose of the ASV and contours of the coastline were obtained and mapped. Specifically in the context of offshore asset inspection, \cite{campos_multi-domain_2020} and \cite{campos_modular_2022} produced a multi-domain map using a multi-beam echosounder as well as a LiDAR, relying on the fusion of GNSS and IMU for odometry. This map includes the environments above and below the surface into a single representation. Building on the above-surface mapping capabilites \cite{jung_consistent_2023} proposed a sensor suite including both optical cameras and RADAR in addition to the LiDAR. Here, the position and heading were likewise obtained by the fusion of GNSS and an Attitude and Heading Reference System. Finally, \cite{rho_high-precision_2025} combined a LiDAR with a DVL and a pressure sensor mounted on an autonomous marine vehicle operating in a structured marine environment. This sensor fusion was used for localization, supporting the mapping process with a forward-looking SONAR.

Unlike traditional crewed patrols, ASVs carrying multimodal sensor platforms can provide continuous, persistent, and cost-effective mapping and surveillance of infrastructure-dense maritime areas, including ports, offshore installations, and cable landing zones, giving unified maps of the marine environment from the seabed to the sky.

Realizing this potential underwater requires careful consideration, as different SONAR types exist, each offering advantages and trades-off. In \cite{campos_multi-domain_2020}, \cite{campos_modular_2022}, and \cite{jung_consistent_2023} multibeam SONARs capable of providing dense 3D point-clouds were used. However, such systems are often expensive and bulky, whereas forward looking SONARS as well as the imaging SONARs employed in \cite{leedekerken_mapping_2014} and \cite{rho_high-precision_2025} are more affordable and compact. This comes at the cost of only producing planar slices of the environment as demonstrated in \Cref{fig:PolarToCartesian}, limiting their ability to capture 3D structures without sufficient motion. One approach to address these limitations is to use multiple smaller SONAR devices. In \cite{joe_sensor_2019} and \cite{joe_sensor_2021}, a horizontally placed FLS and a vertically placed profiling SONAR are used to estimate the elevation of points in the horizontal SONAR image. In \cite{mcconnell_fusing_2020}, \cite{mcconnell_predictive_2021}, and \cite{liu_target_2024} two FLS units are employed in an orthogonal configuration to directly obtain 3D SONAR data in the overlapping beam region at the same rate as the individual SONARs.

While these studies (\cite{leedekerken_mapping_2014}, \cite{campos_multi-domain_2020}, \cite{campos_modular_2022} and \cite{jung_consistent_2023}) all offer mapping capabilities extending to both the above and below surface domains, they all rely on GNSS fused with IMU/AHRS and/or DVL. Therefore, these approaches are not robust to situations where GNSS is either denied due to shadowing from buildings or unreliable due to spoofing. Although SLAM approaches that do not rely on GNSS does certainly exist, such as in \cite{han_coastal_2019}, the generation of comprehensive seabed-to-sky maps under GNSS-denied conditions remains a mostly unexplored and novel research area. 

We present a \emph{GNSS-independent pipeline} for creating 3D \emph{seabed-to-sky} maps—i.e., a single, unified representation spanning both the underwater and above-surface domains. Such a map supports robust navigation and mapping when one domain is feature-sparse and enables multi-vehicle operations that share a common frame in GNSS-denied settings.

\begin{figure}[tbp]
    \centering
    \begin{subfigure}[b]{0.48\columnwidth}
        \centering
        \includegraphics[trim=280 0 280 0, clip, width=\linewidth]{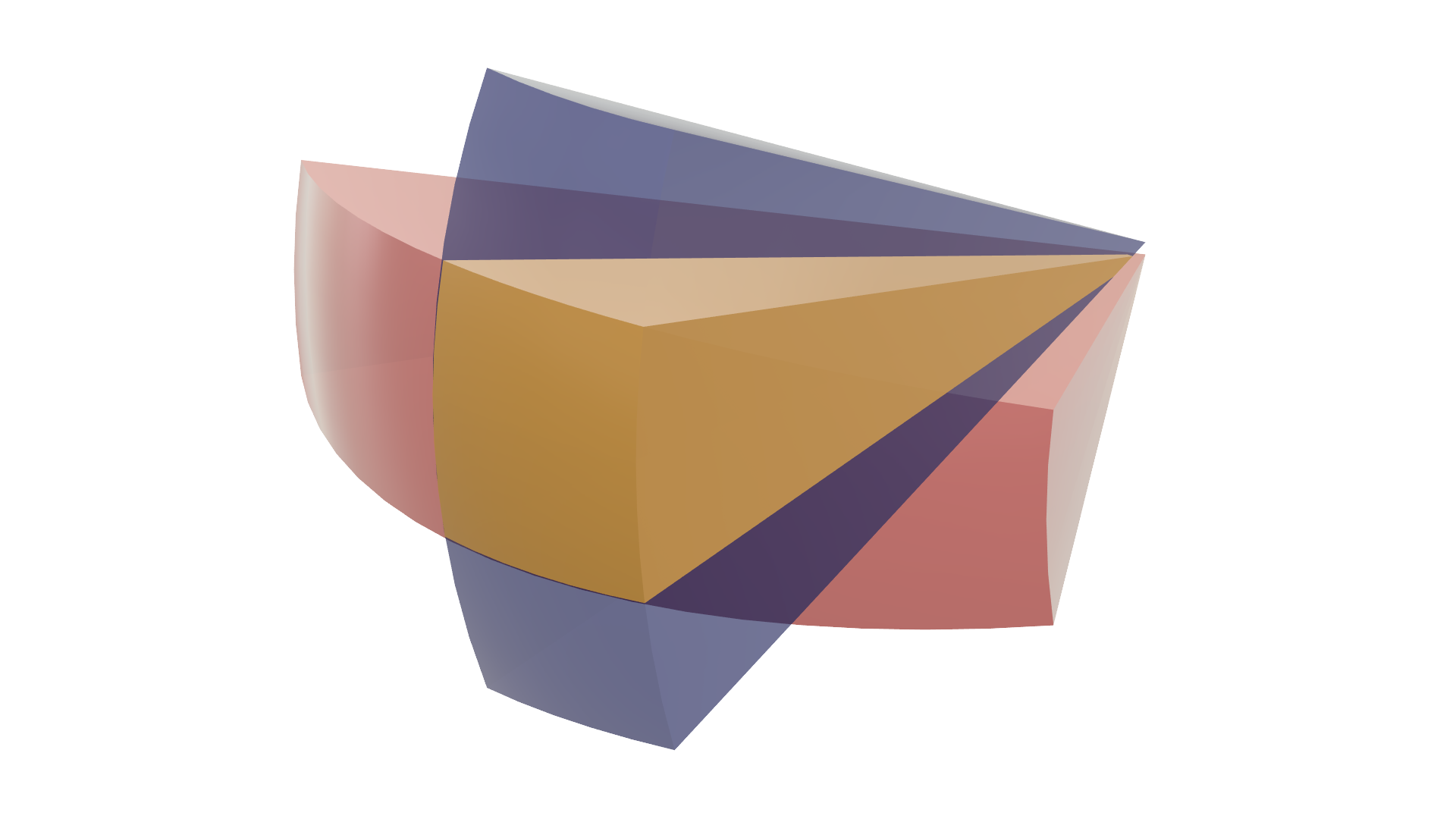}
        \caption{Overlap region of horizontal and vertical SONARs.}
        \label{fig:sonar_overlap}
    \end{subfigure}
    \hfill
    \begin{subfigure}[b]{0.48\columnwidth}
        \centering
        \includegraphics[width=\linewidth]{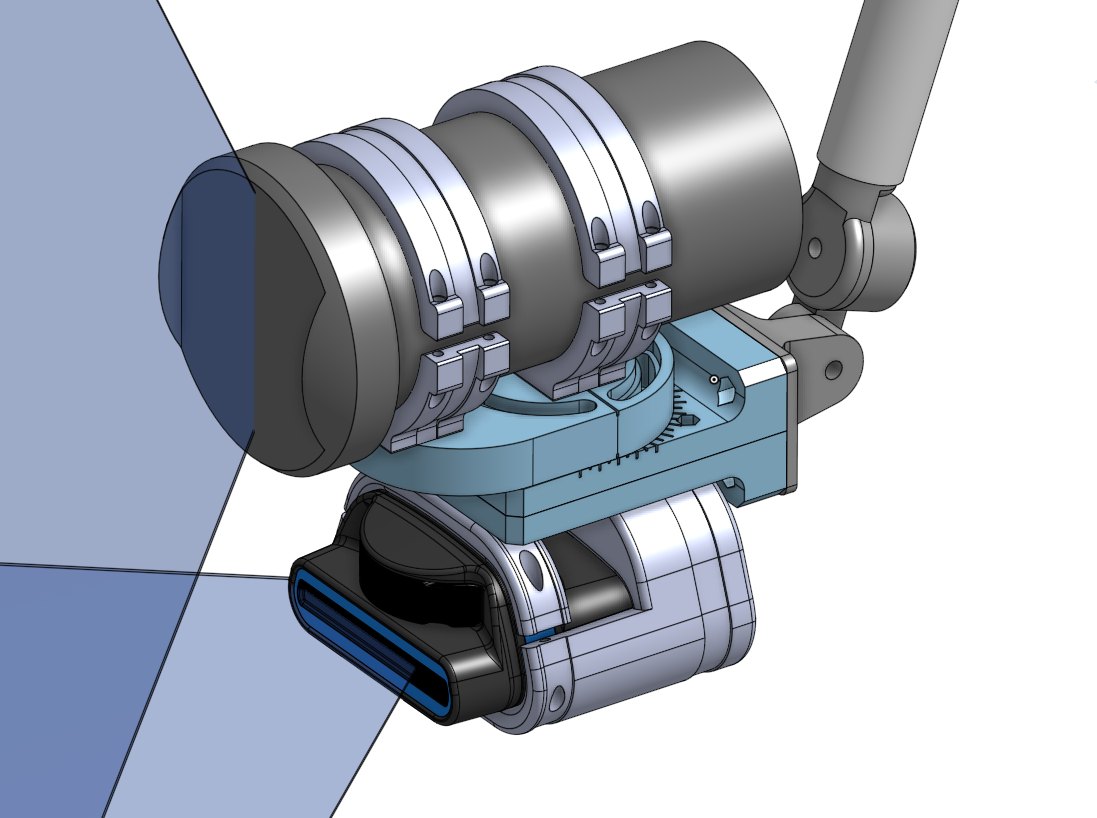}
        \caption{Orthogonal SONAR setup used for 3D reconstruction.}
        \label{fig:orto_setup}
    \end{subfigure}
    \caption{Visualization of the dual SONAR mapping setup. (a) shows the overlap region where 3D point clouds are reconstructed. (b) illustrates the orthogonal SONAR configuration used to capture both horizontal and vertical perspectives.}
    \label{fig:sonar_pipeline}
    \vspace{-4mm} 
\end{figure}

Above the surface, a LiDAR with an internal IMU provides 3D point clouds. Underwater, we adopt the orthogonal stereo method of \cite{mcconnell_fusing_2020} on a custom dual 2D SONAR setup (\Cref{fig:orto_setup}), and \emph{extend it to handle arbitrary inter-SONAR translations} (with the relative rotation fixed to \(90^\circ\)), allowing heterogeneous SONAR models. We also extract a \emph{leading edge} from each SONAR image and inject these as \emph{line-scan} constraints. For the fused seabed-to-sky map, we use LIO-SAM \cite{shan_lio-sam_2020} and \emph{modify it to ingest SONAR stereo points and line-scans both at, and between, keyframes} via motion-interpolated poses.

In summary, the main contributions of the paper are
\begin{enumerate}
    \item A unified seabed-to-sky mapping pipeline for GNSS denied maritime environments.
    \item A dual-SONAR fusion framework, extending \cite{mcconnell_fusing_2020} to support heterogeneous non co-located SONARS.
    \item Integration of both stereo-derived 3D SONAR points with leading-edge line scans for improved coverage in the underwater domain.
    \item Incorporation of acoustic data into the LIO-SAM framework, supporting interpolation between keyframes to integrate sparse SONAR measurements at different update rates.
    \item Real-world validation of the system on an ASV in a marine setting.
\end{enumerate}

\section{Hardware Setup}\label{sec:hardware}
\begin{figure}[tbp]
\centerline{\includegraphics[width=0.6\columnwidth]{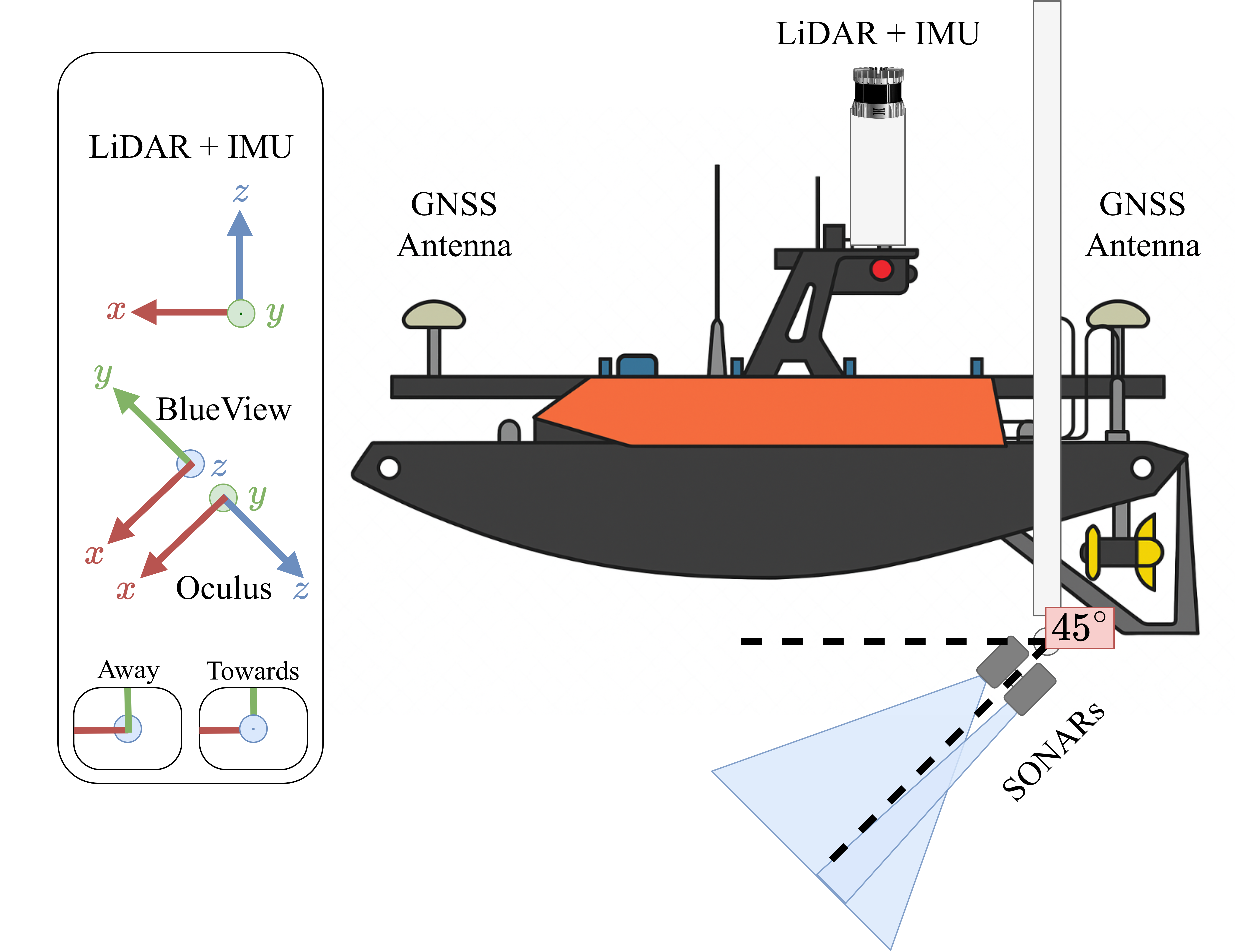}}
\caption{Hardware setup on the Maritime Robotics Otter USV. The dual SONAR setup is pitched $45^{\circ}$ downwards. The reference frames for the sensors are presented in the panel.}
\label{fig:otter_setup}
\end{figure}

The system setup consists of the unmanned surface vehicle \emph{Otter} by Maritime Robotics instrumented with the UBLOX NEO-M8 GNSS system; the Ouster OS-1 LiDAR unit mounted on the top of the vehicle using a custom mount; the stereo Forward Looking Sonar configuration shown in \Cref{fig:orto_setup}. The two SONAR setup is attached to a pole mounted at the stern of the ASV, and the mount is aligned forward relative to the ASV heading, and pitched $45^{\circ}$ downwards. The horizontal SONAR is the Blueprint Subsea Oculus M750d (\emph{Oculus}), and the vertical SONAR is the Teledyne BlueView M900-2250-130-Mk2 (\emph{BlueView}). The technical specifications of the two SONARs are provided in \Cref{tab:sonar-settings}. The LiDAR scans have an update rate of 20 Hz, using full resolution. The IMU has an update rate of 100 Hz. \Cref{fig:otter_setup} shows the complete setup.
\begin{table}[b]
\centering
\caption{Technical specifications of the two SONAR units.}
\begin{tabular}{lcc}
\hline
\textbf{Parameter} & \textbf{Oculus} & \textbf{BlueView} \\
\hline
Ping Rate [Hz]      & 15 & 10 \\
Field of View [°]        & 130 & 45 \\
Operating Frequency [kHz] & 750 & 2250 \\
Gain                     & 80\% & 70 dB \\
Maximum Range [m]        &  10 & 10 \\
Number of Beams          & 512 & 256 \\
\hline
\end{tabular}
\label{tab:sonar-settings}
\end{table}

\subsection{SONAR Imaging}
\begin{figure}[bp]
    \centering
    \includegraphics[width=0.5\columnwidth]{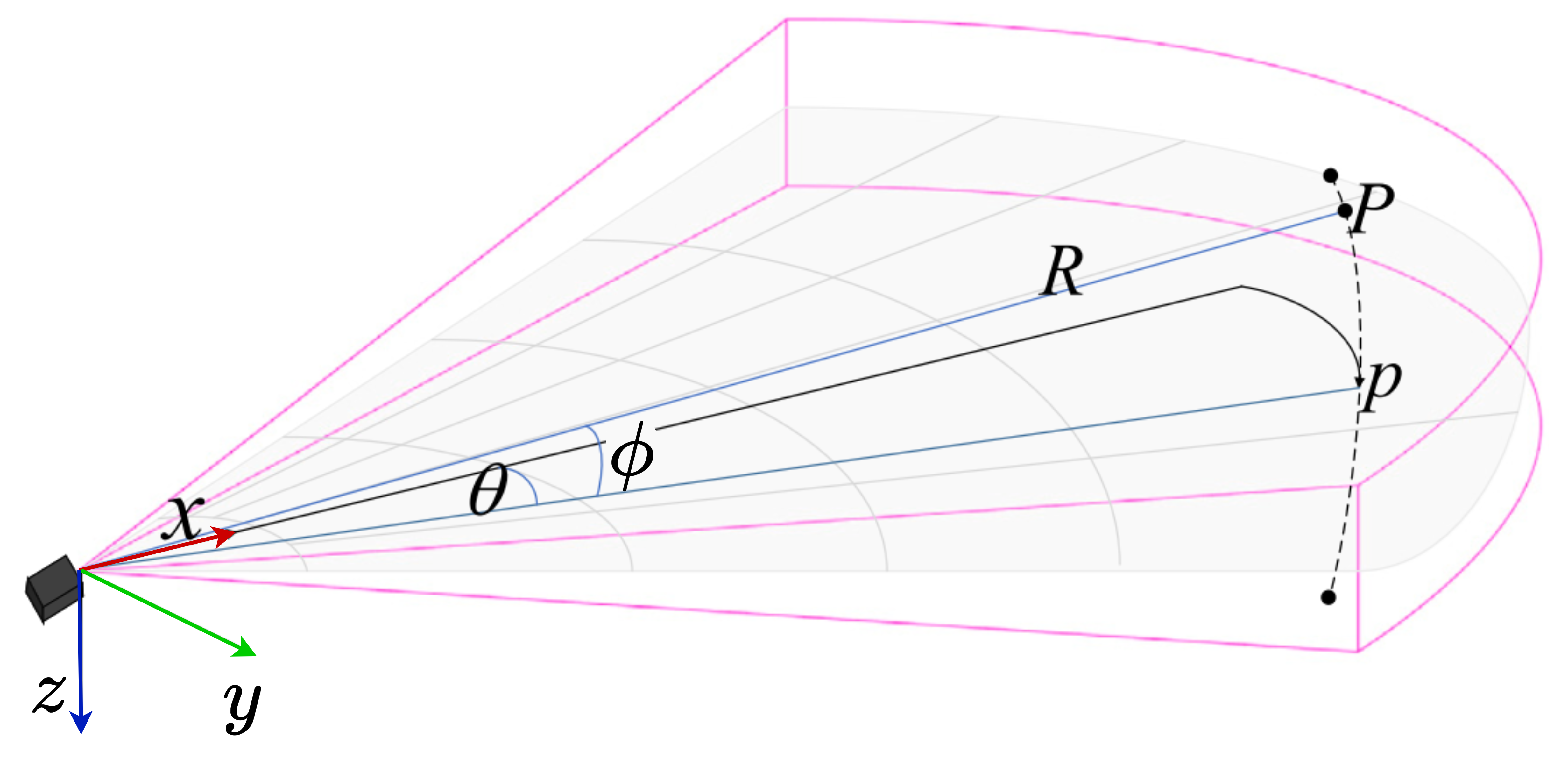}
    \caption{The SONAR projection geometric model. Adapted from \cite{liu_target_2024}, Fig.~7, with minor modifications.}
    \label{fig:PolarToCartesian}
    \vspace{-4mm} 
\end{figure}

Both SONARs used in this work are multi-beam forward looking SONARs, whose data can be represented as a rectangular \textit{beam-range} sonar image $\mathbf{I}(R, \theta) \in \mathbb{R}_+$. A beam direction or \textit{bearing} $\theta \in \Theta \subseteq [-\pi, \pi]$ is represented by each column and a range $R \in \mathbb{R}_{+}$ by each row. Each value in the image represents an intensity with polar coordinates on a 2D plane. Both SONARS in this work project all captured points onto a single plane. \Cref{fig:PolarToCartesian} illustrates this projection, assuming that the elevation $\phi=0^\circ$.
Converting from polar coordinates in the 2D SONAR image to Cartesian 3D coordinates is achieved using the coordinate transformation
\begin{equation}\label{eq:sonar_to_cartesian_plane}
_{\mathcal{S}}\mathbf{p}=
\begin{bmatrix}
x \\
y \\
z
\end{bmatrix}
=
R\begin{bmatrix}
\cos(\theta) \\
\sin(\theta) \\
0
\end{bmatrix}.
\end{equation}

A mapping error is introduced by assuming a point is on the plane as elevation information is lost from the 2D SONAR representation alone depending on the angle, range, and vertical aperture of the setup. This lost elevation information will be reconstructed by fusing the measurements from the overlapping section of the two SONARS in the orthogonal setup, shown as the yellow volume in \Cref{fig:sonar_overlap}. 

\section{Seabed-to-Sky Mapping}
This section describes how the data obtained from the SONARs, LiDAR and IMU are fused in real-time to create the seabed-to-sky map. Raw SONAR images are processed using image processing tools to enable line-scan point-cloud data extraction in the projected SONAR plane via \textit{leading-edge detection}. Features in the overlapping SONAR views are matched to recover the 3D structure, while the LiDAR and IMU data are integrated using a mapping algorithm, for localization and above-water mapping. The SONAR data is then appended to the map, producing the unified representation of data above and below the water surface.
\begin{figure}[tbp]
    \centering
    \includegraphics[width=0.9\columnwidth]{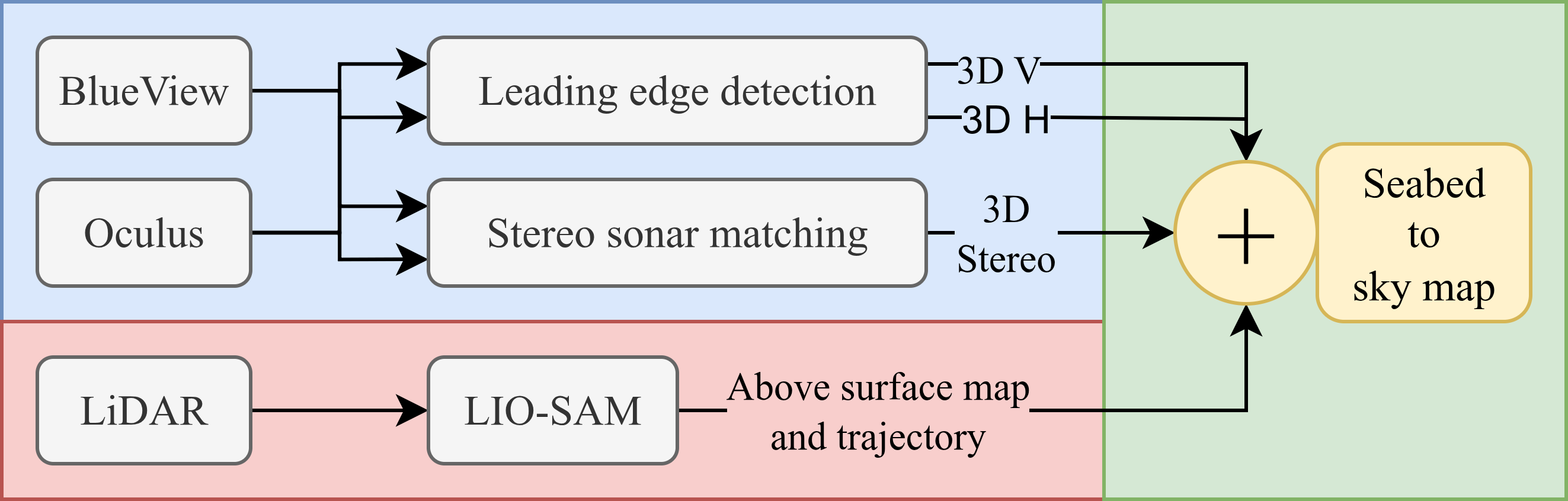}
    \caption{The full mapping pipeline combining both SONARs with the LiDAR+IMU. Leading-edge detection and stereo matching from the SONARs are integrated, and the resulting 3D point cloud is added to the LiDAR map.}
    \label{fig:pipeline_full}
    \vspace{-4mm} 
\end{figure}

\subsection{Sonar Image Pre-Processing}\label{sec:subsec_image_processing}
To deal with noise and artefacts in the SONAR images, and enable robust feature extraction, an image preprocessing step is utilized. The convention for describing SONAR images in the polar bearing-range representation outlined above is used. Noteworthy that these preprocessing steps are essential for the SONAR data used in this work. The use of different SONARs with other sensing settings may require alternative pre-processing methods. 

\noindent \emph{Horizontal SONAR.} To rectify noise within the \emph{Oculus} images, a four-step pre-processing is applied. First, the $10\%$ row quantile is subtracted to account for intensity variations across row groupings and to reduce the amount of low-intensity values. Next, the Otsu's method \cite{otsu_threshold_1979} is applied to create a mask that keeps the relevant high intensity values. The resulting image is then normalized and converted to 8-bit format, for consistent intensity scale between the two SONARs. Finally, a $3\times1$ open-operation, further lowers the amount of noise. This kernel is chosen as the noise spans over multiple columns (bearings), making a tall kernel suitable. Larger kernels sizes were found to make the edge``bleed'' into the surface. An example of a processed image is shown in \Cref{fig:sonar_pipeline}.
\begin{figure}[tbp]
    \centering
    \includegraphics[width=0.7\columnwidth]{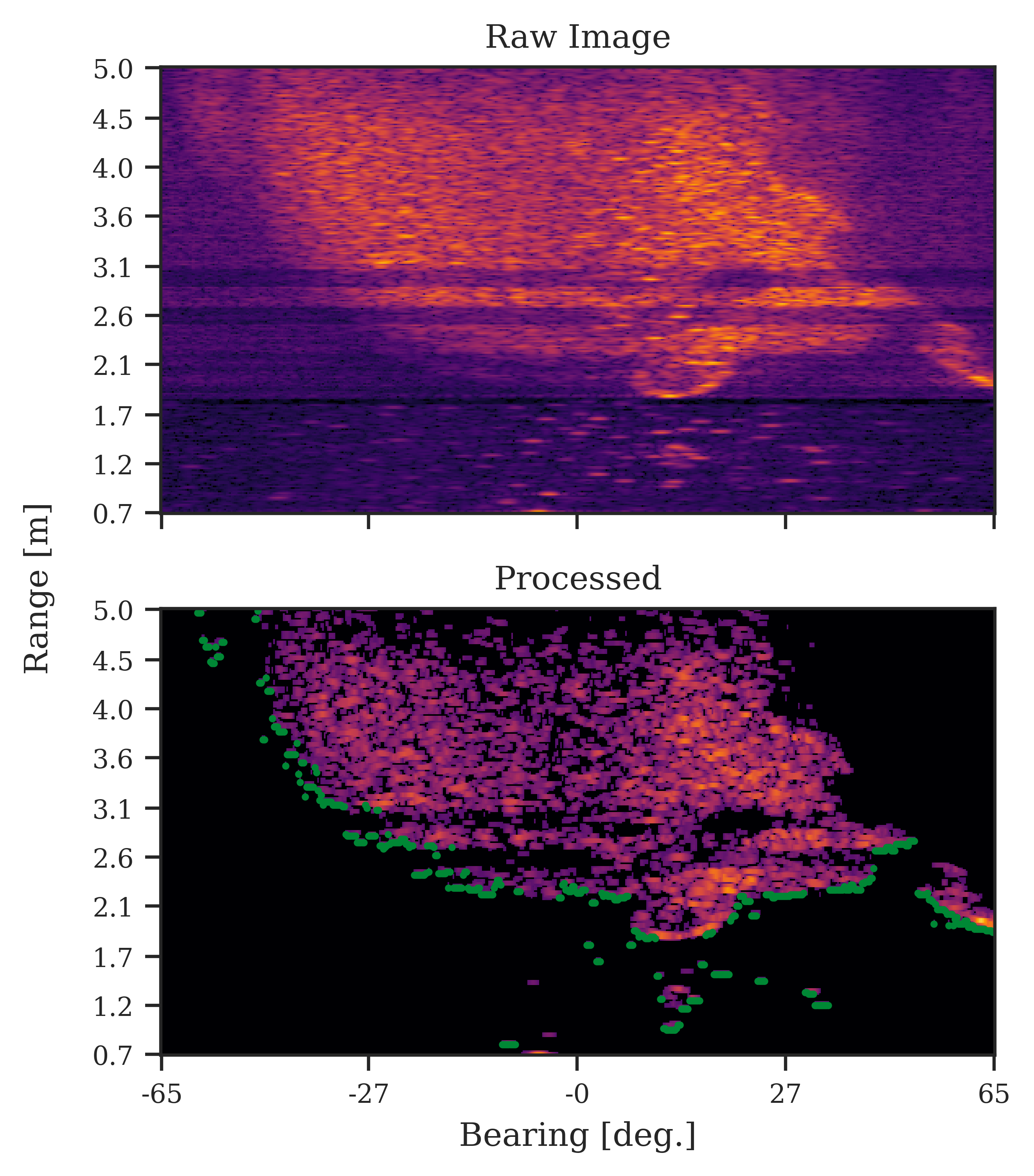}
    \caption{An \emph{Oculus} image before and after pre-processing. The leading edge is highlighted in green.}
    \label{fig:sonar_pipeline}
    \vspace{-4mm} 
\end{figure}

\begin{figure}[tbp]
    \centering
    \includegraphics[width=0.7\columnwidth]{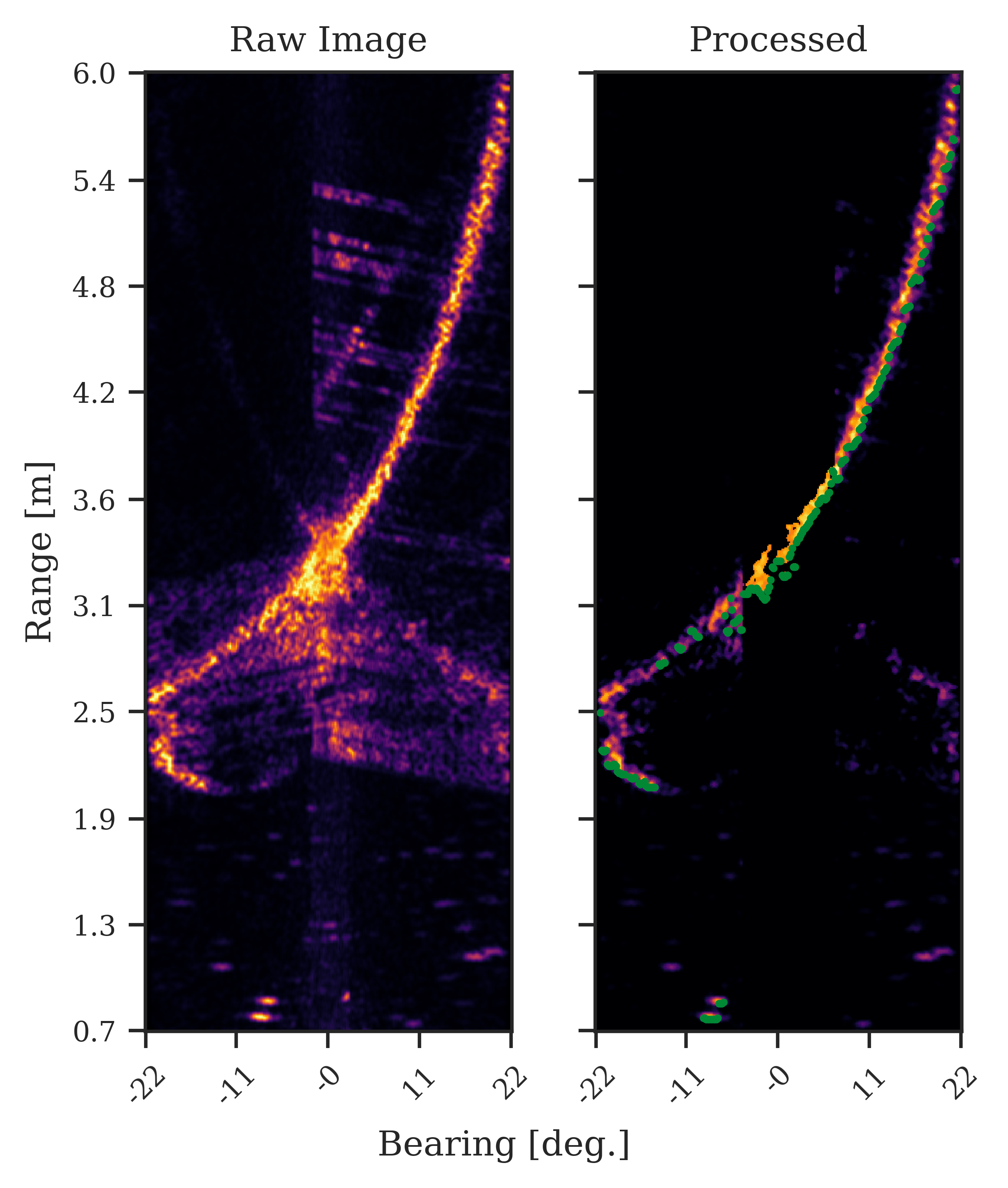}
    \caption{A \emph{BlueView} image before and after pre-processing. The leading edge is highlighted in green.}
    \label{fig:sonar_vertical_pipeline}
    \vspace{-4mm}
\end{figure}

\noindent \emph{Vertical SONAR.} For the \emph{BlueView}, high-gain artifacts are rectified using a similar procedure. Row-wise mean subtraction is performed in order to reduce artifacts induced across whole rows. Low-level intensity pixels are removed from the centre of the image between bearings $\theta\in[-10^{\circ}, 10^{\circ}]$, as artefacts are prominent in this region. The image is then normalized and converted to an 8-bit format, and a $3\times3$ median filter reduces the remaining noise. An example of a processed image is shown in \Cref{fig:sonar_vertical_pipeline}.

\subsection{Leading Edge Detection}
To supplement the stereo SONAR approach that only uses the overlapping section of the SONAR images, line-scans spanning the full sonar field of view are extracted. To obtain this information, a \textit{leading edge} method is utilized. Here, it is assumed that all points on the SONAR image are on a single plane neglecting the uncertainty introduced in the elevation dimension. 

Given a rectangular SONAR image $\mathbf{I}(R, \theta)$, for each beam direction $\theta$ the leading edge is determined by identifying the smallest range $R$ at which the image intensity exceeds a given intensity threshold. This is expressed as the function $r(\theta)$, associating each beam direction $\theta$ with the corresponding range $R$:
\begin{equation}
r(\theta)=\min\{R|\mathbf{I}(\theta,R)>\tau\},\label{eq:leading_edge}\end{equation}
where $r(\theta)$ is the range corresponding to the leading edge in beam direction $\theta$, $\mathbf{I}(\theta, R)$ represents the magnitude of the sonar at range $R$ and beam direction $\theta$, and $\tau$ is either a predefined or adaptive threshold value. In this work, this threshold is set  to $80$ for the \emph{Oculus} and $130$ for the \emph{BlueView} (the maximum intensity in 8-bit image is $255$). 

The leading edges for a pair of SONAR images are depicted with green points in \Cref{fig:sonar_pipeline}.

\subsection{Data Association}
Although the leading edge provides a range estimate along each beam of the SONAR, the projection in \Cref{eq:sonar_to_cartesian_plane} (illustrated in \Cref{fig:PolarToCartesian}) does not capture variations in elevation. The elevating dimension can be reconstructed by matching points between the horizontal and vertical SONARs, associating the data between them. This allows for the construction of a 3D point cloud in the area where the vertical apertures of both SONARs overlap, as seen in \Cref{fig:sonar_overlap}. In this work, each SONAR has a $20^{\circ}$ vertical aperture, resulting in a $20\times20$ window for matching. The implementation of this step of the map generation is inspired by \cite{mcconnell_fusing_2020} building on their notation, with the following key modification:
In the original work, polar coordinates are used throughout. This introduces a hard constraint in the physical setup, i.e. the SONAR heads need to be perfectly aligned in the horizontal plane. This challenges the use of SONARS from different makers, which have different phase centres, as is the case for this implementation\footnote{In \cite{mcconnell_fusing_2020} and \cite{mcconnell_predictive_2021}, the same brand and model of sonars was used.}. Moreover, the vertical displacement of the SONAR heads introduces additional complications as the correction to account for this displacement is, in the original implementation, applied after associating data in the two SONAR images, which might lead to erroneous data association when matching features, as matching relies in part on the vertical axis. To mitigate these issues, we extend the method \cite{mcconnell_fusing_2020} by utilizing both polar \textit{and} Cartesian coordinate representations, thus allowing for displaced sonar configurations. Additionally, the de-noising step in the original work is replaced by the pre-processing pipeline above, to accommodate the choice of SONAR models and settings.

Following this change from a polar to Cartesian coordinate system, a SONAR image is represented as $xyz$ intensity vectors:
\begin{equation}\mathbf{z}^{(h)}_h=[x^{(h)}_hy^{(h)}_hz^{(h)}_h\gamma^{(h)}_h]^\top,\:\mathbf{z}^{(v)}_v=[x^{(v)}_vy^{(v)}_vz^{(v)}_v\gamma^{(v)}_v]^\top,\label{eq:intensity_vectors}
\end{equation}
where the subscript indicates either the horizontal $h$ or vertical $v$ coordinate system, the superscript denotes the data source i.e. horizontal $(h)$ or vertical $(v)$ SONAR, and $\gamma$ representing the intensity. For the presented method, the inertial frame of the horizontal SONAR will be used as the reference base-frame. The relation between the reference frames of the horizontal and vertical SONAR can be seen in \Cref{fig:otter_setup}. The points from the horizontal SONAR can be expressed in the horizontal base-frame
\begin{equation}
\begin{bmatrix}x^{(h)}_h\\y^{(h)}_h\\z^{(h)}_h\end{bmatrix}=R^{(h)}_h\begin{bmatrix}\cos(\theta^{(h)}_h)\\\sin(\theta^{(h)}_h)\\0\end{bmatrix}.
\end{equation}
For the vertical SONAR the points are transformed into the horizontal SONAR frame using homogeneous coordinates:
\begin{equation}
\setlength{\arraycolsep}{2pt}
\begin{bmatrix}x^{(v)}_h\\y^{(v)}_h\\z^{(v)}_h\\1\end{bmatrix}=
\begin{bmatrix}
\mathbf{R}_{3\times3} & \mathbf{0}_{3\times1} \\
\mathbf{0}_{1\times3} & 1
\end{bmatrix}
\begin{bmatrix}
\mathbf{I}_{3\times3} & \mathbf{t}_{3\times1} \\
\mathbf{0}_{1\times3} & 1
\end{bmatrix}\begin{bmatrix}R^{(v)}_v\cos(\theta^{(v)}_v)\\R^{(v)}_v\sin(\theta^{(v)}_v)\\0\\1\end{bmatrix},\label{eq:vh_trans}\end{equation}
where $\mathbf t$ is the translation vector, $\mathbf R$ is the rotation matrix (in this work $-90^{\circ}$ around the $x$-axis) and $\mathbf I$ represents the identity matrix. From this point forward, all coordinates are expressed in the horizontal SONAR frame, and the subscript indicating coordinate frame will be omitted for simplicity. Given that each image contains $N\in\mathbb{N}$ intensity vector observations, the data from both SONAR images can be expressed as sets
\begin{equation}Z^{(h)}=\{\mathbf{z}_1^{(h)},\cdots,\mathbf{z}_N^{(h)}\},\quad Z^{(v)}=\{\mathbf{z}_1^{(v)},\cdots,\mathbf{z}_N^{(v)}\}.\end{equation}
The data association problem can be considered a vertex matching problem in a bipartite graph, where all observed intensity points are vertices $\mathcal V=Z^{(h)}\cup Z^{(v)}$, and the set of all possible associations for a data point are interpreted as an edge:
\begin{equation}
    \mathcal E_i=\{(\mathbf z_i^{(h)}, \mathbf z_1^{(v)}),...,(\mathbf z_i^{(h)}, \mathbf z_N^{(v)})\}.    
\end{equation}
The union of all edge sets gives the total edge set $\mathcal E_i=\cup_{i=1}^{N}\mathcal E_i$. Hence, the solution can be found by posing an optimization problem by defining a loss between features $\mathcal{L}(\mathbf z_i^{(h)}, \mathbf z_j^{(v)})$, obtaining solutions such that
\begin{equation}
S = \bigcup_{\mathcal E_i \in \mathcal E}\quad 
    \underset{(\mathbf{z}_i^{(h)}, \mathbf{z}_j^{(v)}) \in \mathcal E_i}{\text{argmin}} 
    \mathcal{L}(\mathbf{z}_i^{(h)}, \mathbf{z}_j^{(v)}),
    \quad \mathrm{where}\; S \subset \mathcal E.\label{eq:optimization_problem}
\end{equation}
A bijective solution is required to prevent matching duplications in the set of solutions $S$. After obtaining this set of unique solutions, information from both sensors can be fused into Cartesian coordinates using the average of the extracted coordinates from the two sensors:
\begin{equation}
\mathbf{p}=\begin{bmatrix}x\\y\\z\end{bmatrix}=\frac{1}{2}\begin{bmatrix}x^{(h)}+x^{(v)}\\y^{(h)}+y^{(v)}\\z^{(h)}+z^{(v)}\end{bmatrix}.\label{eq:cartesian_coordinates}
\end{equation}
\begin{figure*}[t]
    \centering
    \includegraphics[width=\textwidth]{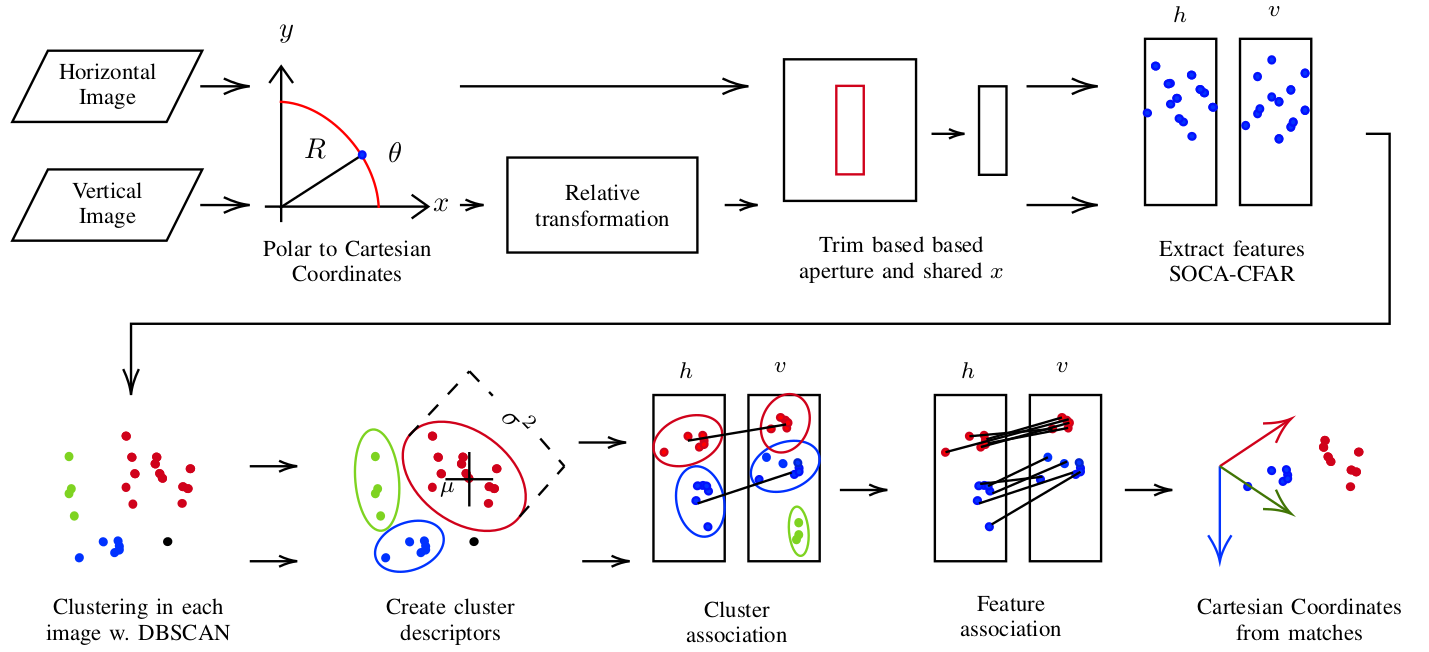}
    \caption{Conceptual diagram of the proposed method. The input images are already pre-processed, according to \Cref{sec:subsec_image_processing}. After being transformed into Cartesian coordinates, a transformation to the horizontal SONAR frame is performed. The images are trimmed based on intersecting areas, reducing processing time in the following steps. Features are extracted from each image and clustered. These clusters are matched based on their descriptors. Features are then matched between the two images, restricted by the assigned cluster pair. The matched features are at last converted into 3D points in Cartesian space.}
    \label{fig:pipeline}
\end{figure*}
\vspace{-1mm}

\subsection{Point Cloud Generation}
Having specified a generalized framework for data association between the two SONAR images, the complete method for generating a 3D point cloud in Cartesian coordinates can be presented, solving the optimization problem posed in \Cref{eq:optimization_problem}. \Cref{fig:dual-fls-diagram} shows a complete conceptual overview of the method. After converting to Cartesian coordinates and transforming the vertical SONAR data into the horizontal frame, the images are trimmed to their overlap to reduce feature extraction time. 

For feature extraction, the Smallest of Cell Averages Constant False Alarm Rate (SOCA-CFAR) is applied. To improve robustness, clustering and association between clusters is performed to take advantage of the underlying surfaces that the features represent. Density-Based Spatial Clustering of Applications with Noise (DBSCAN) is used to cluster the features, identifying arbitrary clusters without prior knowledge of the SONAR image. This extraction and clustering method follows \cite{mcconnell_fusing_2020}, with the hyperparameters for SOCA-CFAR shown in \Cref{tab:cfar_settings}. For DBSCAN, two hyperparameters are set: $\epsilon=0.20$, the maximum distance to a nearby candidate, and $n=20$, the minimum samples per cluster. For both methods, hyperparameters are obtained through iterative tuning on SONAR images selected from the data.

\begin{table}[tbp]
    \centering
    \caption{CFAR settings for Horizontal and Vertical Sonars. The minimum intensity is used to set a lower threshold of intensity that is allowed for the cell to be considered a feature.}
    \label{tab:cfar_settings}
    \begin{tabular}{lcc}
        \toprule
        Parameter & Horizontal Sonar & Vertical Sonar \\
        \midrule
        Reference cells $N_{rc}$ & 16 & 24 \\
        Guard cells $N_{gc}$ & 8 & 8 \\
        Probability of False Alarm $P_{\text{fa}}$ & 0.2 & 0.2 \\
        Minimum Intensity & 100 & 130 \\
        \bottomrule
    \end{tabular}
\end{table}

After defining feature clusters for each SONAR image, descriptors $\mathbf{c_t}$ are specified to associate clusters across the two SONARS. In the Cartesian coordinate space, these descriptors are based on the position of the cluster in the $x$ coordinate direction
\begin{equation}
\mathbf{c}_t=\begin{bmatrix}\mu&\sigma^2&x_{\min}&x_{\max}\end{bmatrix},
\end{equation}
where $\mu$ and $\sigma^2$ are mean and variance of the cluster in the $x$ coordinate direction.
Each cluster $\mathbf{c}_t^{(h)}$ is a assigned to the cluster $\mathbf{c}_t^{(v)}$ from the vertical SONAR
that minimizes the cost function
\begin{equation}
\mathcal{L}(\mathbf{c}_t^{(h)},\mathbf{c}_t^{(v)})=||\mathbf{c}_t^{(h)}-\mathbf{c}_t^{(v)}||_2.
\end{equation}

Following the definition and association of clusters of interest across the two SONARS, feature association is performed in a similar way within each pair of clusters. Descriptors include the feature’s $x$-coordinate and intensity, as well as local intensity context, based on the assumption that similar features will have similar surrounding intensities. Specifically, $\bar{\gamma_x}$ and $\bar{\gamma_y}$ represent the mean of the feature intensity $\gamma$ with its immediate neighbours along the $x$- and $y$-axes in the Cartesian SONAR image. In \cite{mcconnell_fusing_2020} orthogonality is leveraged by switching the directions for the vertical SONAR, which is also the case here:
\begin{align}
    \mathbf{f}(\mathbf{z}_i^{(h)}) &= \begin{bmatrix}x&\gamma&\bar{\gamma_x}&\bar{\gamma_y}\end{bmatrix}^\top,\\
    \mathbf{f}(\mathbf{z}_j^{(v)}) &= \begin{bmatrix}x&\gamma&\bar{\gamma_y}&\bar{\gamma_x}\end{bmatrix}^\top.
\end{align}

Creating the mapping $\mathbf{f}$ from features $\mathbf{z}$ to feature descriptors $\mathbf{f}(\mathbf{z})$, feature association is carried out within each of the associated pairs of clusters from the previous step, such that the cost function is minimized:
\begin{equation}
\mathcal{L}(\mathbf{z}_i^{(h)},\mathbf{z}_j^{(v)})=||\mathbf{f}(\mathbf{z}_i^{(h)})-\mathbf{f}(\mathbf{z}_j^{(v)})||_2.\label{eq:feature_cost_function}
\end{equation}
Finally with this set of solutions, the Cartesian coordinates for each point in the 3D point-cloud are calculated using \Cref{eq:cartesian_coordinates}.

\subsection{Map Generation}
For creating the seabed-to-sky map itself, including both the LiDAR and SONAR point clouds, the LIO-SAM method \cite{shan_lio-sam_2020} was chosen. LIO-SAM addresses several issues with previous LiDAR-based SLAM approaches, including drifting induced by raw skewed LiDAR measurements; inefficient scan matching; lack of tightly-coupled LiDAR-IMU systems and problems with real-time performance. LIO-SAM is modelling the state estimation problem as a factor graph with a Gaussian Noise Model, therefore it can be solved as a nonlinear least-squares problem\cite{shan_lio-sam_2020}\cite{dellaert_factor_2017}. Although the implementation of the algorithm has been extended by adding the SONAR point cloud and leading-edge linescans in keyframes, only a short account of it is given.

Using LIO-SAM for both robust localization and mapping, dense 3D ground maps can be constructed by using the extracted features from the LiDAR measurements, associated with each node in the factor graph. To extend this framework with acoustic sensing, the SONAR point clouds are incorporated into the map by associating them with the temporally corresponding LiDAR keyframes. This maintains the efficient graph representation allowing for large updates due to loop closures. In this work, emphasis is on the association of the SONAR point clouds to the estimated nodes to create a map, while excluding them from the graph optimization itself.

\begin{figure}[tbp]
\centerline{\includegraphics[width=\columnwidth]{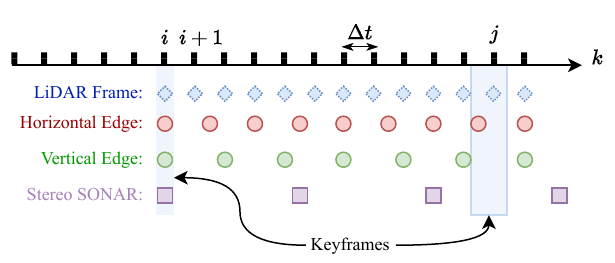}}
\caption{Interpolation between the LiDAR keyframes, showing the integration of both leading-edge and the stereo SONAR method with varying update rates into the map.}
\label{fig:keyframes}
\vspace{-4mm}
\end{figure}

For the SONAR data obtained in-between the LiDAR keyframes, both from the leading-edge process and the stereo matching, positions are interpolated under the assumption of constant velocity and linear motion. This is implemented as data obtained from the SONARS is much sparser than that obtained form the LiDAR. Keyframes are inserted whenever the ASV translates by approximately \(1\,\mathrm{m}\) or rotates by \(2.86^{\circ}\) (\(\approx 0.05\) rad), ensuring coverage without redundant updates. This interpolation lets us fuse sensors with different update rates rather than restricting updates to keyframes (see \Cref{fig:keyframes}). The resulting seabed-to-sky map integrates horizontal and vertical leading-edge data, stereo-derived 3D points, and the above-surface LiDAR map. The map updates at \(\sim 2.65\,\mathrm{Hz}\), while odometry and trajectory estimates update at \(\sim 2.85\,\mathrm{Hz}\).


\section{Results}
For the evaluation of the proposed mapping pipeline, RTK reference points are used. The points have an average standard deviation of 0.011 m and are given in the European Terrestrial Reference System 1989 (ETRS89) UTM32. As the obtained seabed-to-sky map is given in a local reference frame with an arbitrary heading, the rotation and translation to the East-North-Up frame in which the RTK points are defined must be estimated. This is achieved using Singular Value Decomposition, visually selecting points in the map corresponding to the RTK points. Having applied the obtained translation and rotation to the selected points, the error is computed using the Euclidean norm. The presented results are based on an experimental campaign where the \emph{Otter} was deployed in Belvederekanalen (Copenhagen, Denmark), on the 5th of December 2024.

\begin{figure}[tbp]
    \centering
    \begin{subfigure}[b]{0.49\columnwidth}
        \centering
        \includegraphics[width=\linewidth]{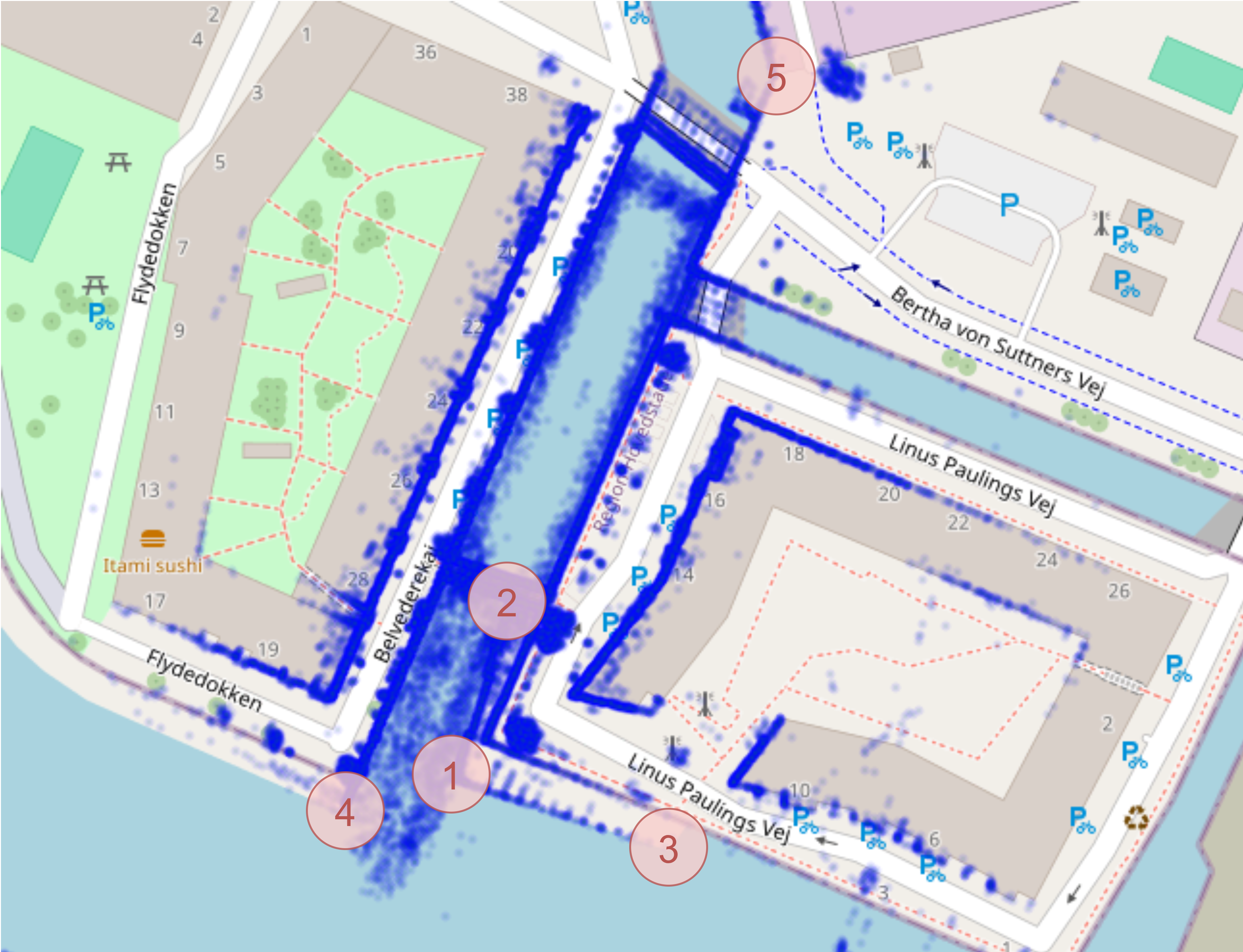}
        \caption{The constructed seabed-to-sky map (Experiment 4) plotted on top of a satellite image of Belvederekanalen. The red circles indicate the RTK reference points used for evaluation.}
        \label{fig:rtk_on_map}
    \end{subfigure}
    \hfill
    \begin{subfigure}[b]{0.49\columnwidth}
        \centering
        \includegraphics[width=\linewidth]{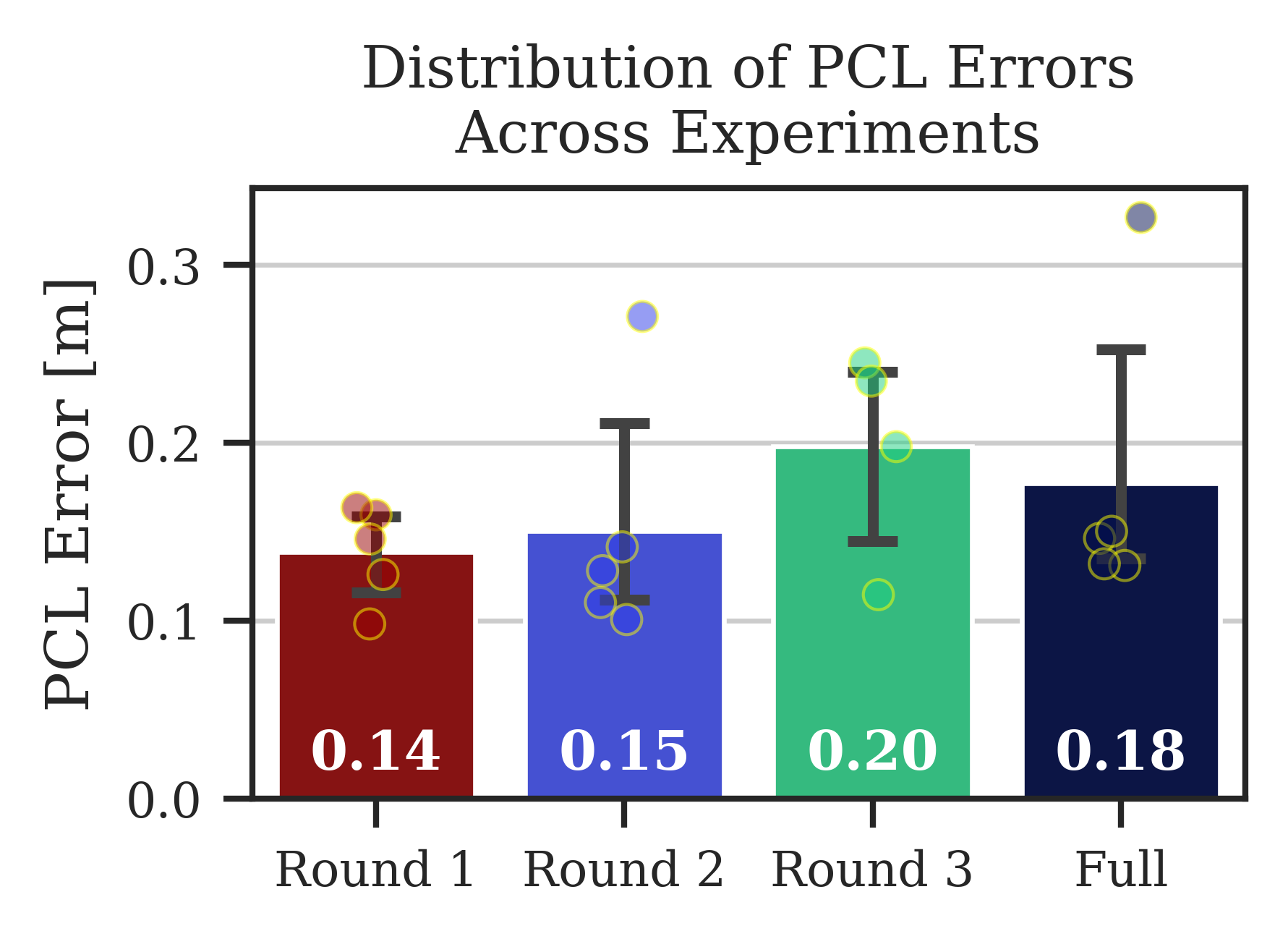}
        \caption{Error given as the Euclidean distance between RTK points and the corresponding points picked from the constructed ground map. Error bars indicate the 95\% confidence interval.}
        \label{fig:pcl_errors}
    \end{subfigure}
    \caption{Visualization and evaluation of seabed-to-sky mapping using RTK reference points.}
    \label{fig:rtk_eval}
    \vspace{-4mm}
\end{figure}

The following experiments were performed:
\begin{enumerate}
    \item Clockwise survey no. 1 in the canal.
    \item Clockwise survey no. 2 in the canal.
    \item Lawnmower pattern survey at the mouth of the canal.
    \item Continuous run over the previous three rounds.
\end{enumerate}
Data from each experiment is separately processed, and thus four maps are generated, and shall be evaluated. Five RTK reference points were chosen, and they are shown in \Cref{fig:rtk_on_map}.

In round 3, point 5 is excluded as the USV never passed the first bridge. \Cref{fig:rtk_on_map} and \Cref{fig:rtk_eval} shows the average and $95\%$ confidence interval of each round. The method yields consistent results, with mean errors in the range $0.139--0.198$ m across all four runs, and confidence intervals overlapping.

While the RTK points serve to evaluate the above-surface map obtained with the LiDAR, the seabed maps must be evaluated with other means, due to a lack of underwater reference points. 
A quantitative evaluation of SONAR reconstruction of the canal walls can be made as this structure is present also on the map generated by the LiDAR. A 40 m long segment of the easternmost wall of the canal is chosen to carry out the analysis. The map is centered and rotated such that the wall aligns with the $x$-axis in a Cartesian coordinate system, with the $z$-axis pointing upwards. All LiDAR points that are below the water surface are disregarded due to noise and inaccuracy caused by reflection. To mitigate noise, the 5\% lowest and highest values in the $y$-dimension are removed.

To evaluate the similarity between the LiDAR and dual-SONAR point distributions, four metrics are employed: the estimated wall width in the $y$-direction, the mean pairwise cosine similarity across all point correspondences, the Kernel Density Estimate (KDE) along the wall, and the Hellinger distance between the resulting distributions. The results are shown in \Cref{tab:results}.

\begin{table}[tbp]
    \centering
    \caption{SONAR and LiDAR point clouds alignment.}
    \label{tab:results}
    \begin{tabular}{lccc}
        \toprule
        Metric & Exp. 1 & Exp. 2 & Exp. 4 \\
        \midrule
        \makecell[l]{SONAR--LiDAR estimated wall \\ width difference (m)} & -0.188 & -0.168 & -0.230 \\
        Avg. Cosine Similarity & 0.993  & 0.991  & 0.993 \\
        Hellinger Distance & 0.46 & 0.48 & 0.44 \\
        \bottomrule
    \end{tabular}
    \vspace{-4mm}
\end{table}

\begin{figure}[bp]
    \centering

    \begin{subfigure}[b]{0.48\linewidth}
        \centering
        \includegraphics[width=\linewidth]{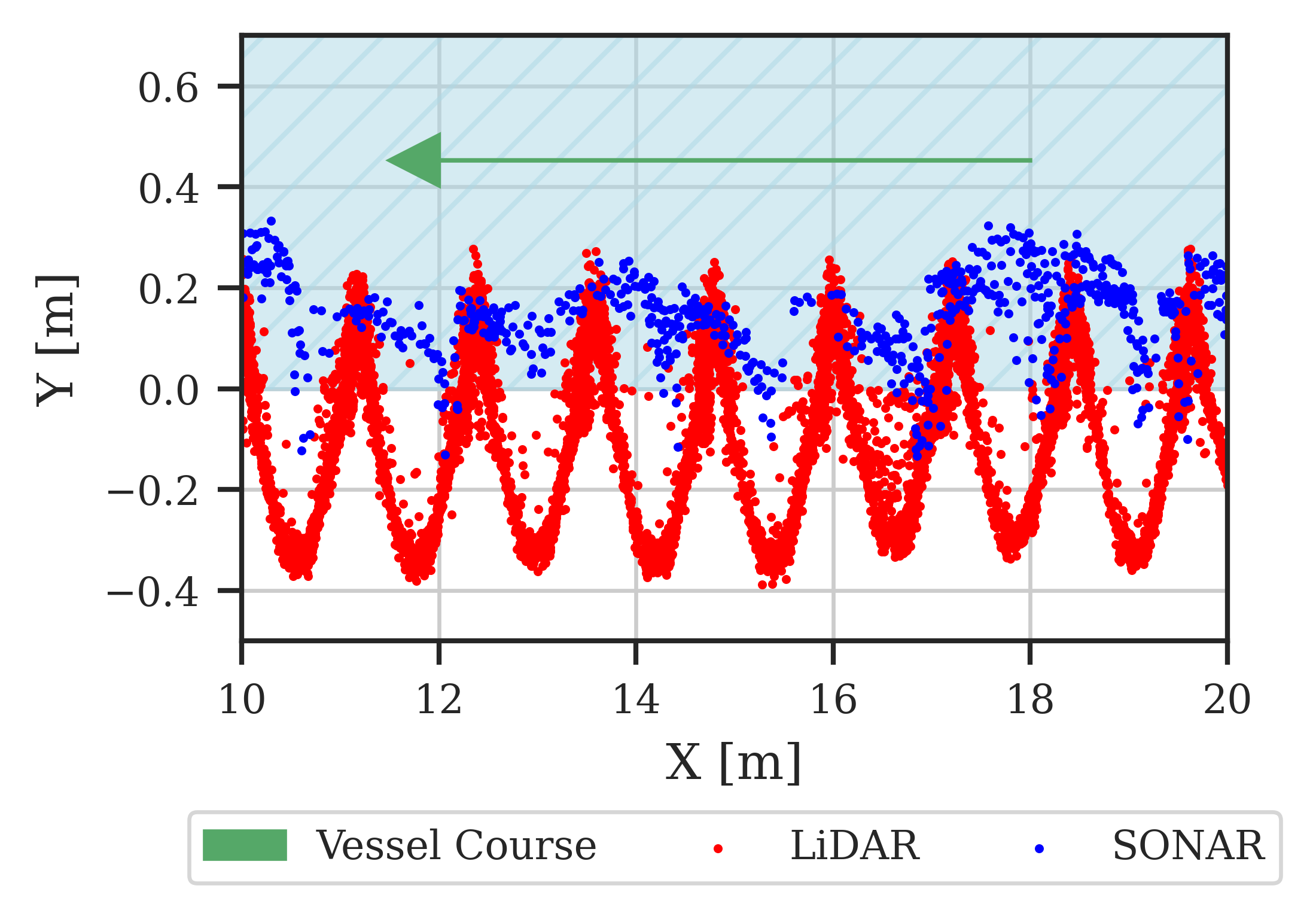}
        \caption{Point-clouds wall. Round 1.}
        \label{fig:wall1_round1}
    \end{subfigure}
    \hfill
    \begin{subfigure}[b]{0.48\linewidth}
        \centering
        \includegraphics[width=\linewidth]{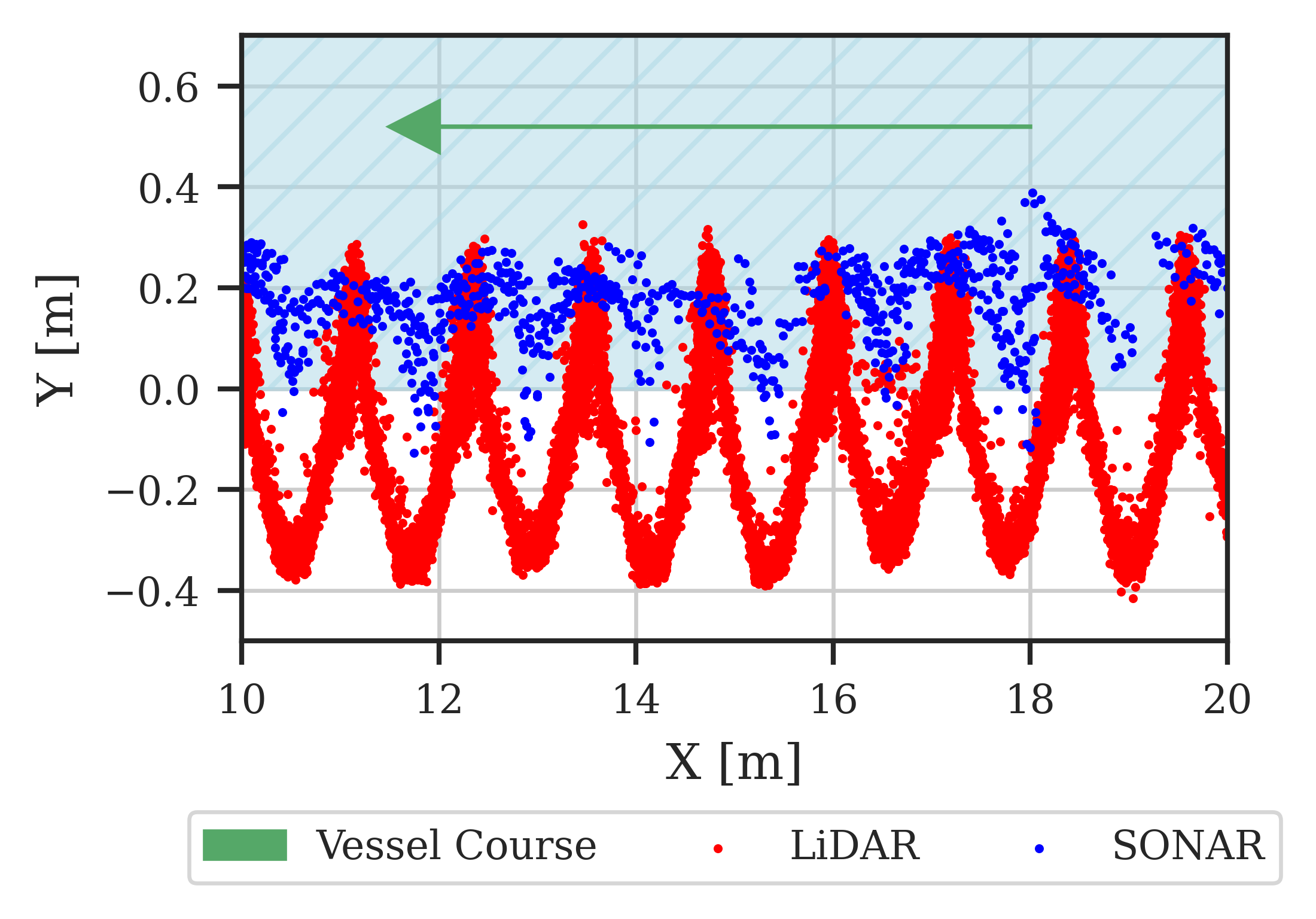}
        \caption{Point-clouds wall. Full.}
        \label{fig:wall1_full}
    \end{subfigure}

    \vspace{2mm} 

    \begin{subfigure}[b]{0.48\linewidth}
        \centering
        \includegraphics[width=\linewidth]{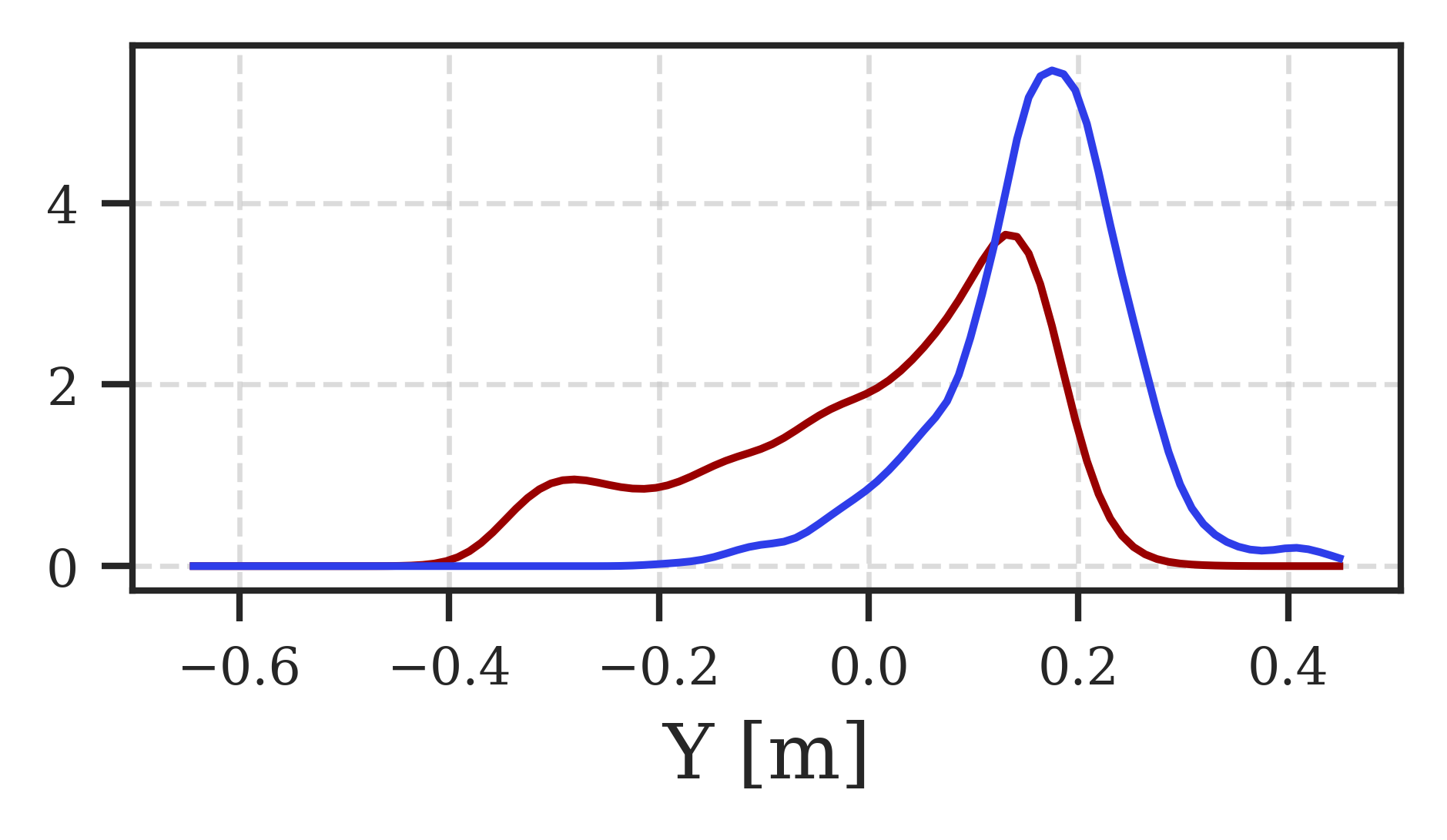}
        \caption{KDE for wall in Round 1. Hellinger distance: 0.46.}
        \label{fig:kde_wall1_round1}
    \end{subfigure}
    \hfill
    \begin{subfigure}[b]{0.48\linewidth}
        \centering
        \includegraphics[width=\linewidth]{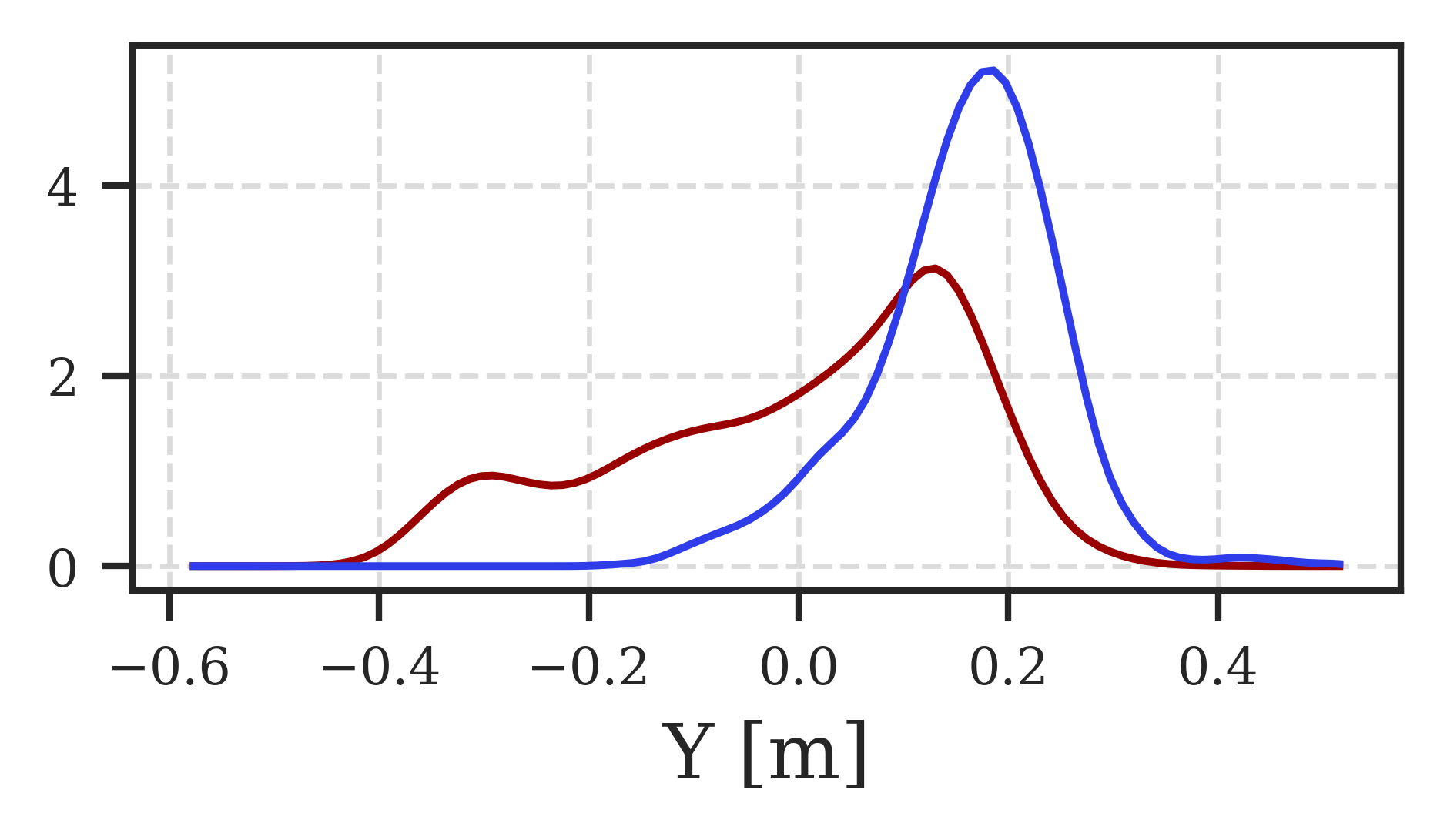}
        \caption{KDE for wall in full run. Hellinger distance: 0.44.}
        \label{fig:kde_wall1_full}
    \end{subfigure}

    \caption{Reconstruction efforts of selected wall for Experiments 1 and 4.}
    \label{fig:interpolation_wall}
\end{figure}

\Cref{tab:results} shows that the SONAR tends to consistently underestimate the width of the wall compared to the LiDAR. However, it should be noted that the LiDAR map is not a ground truth and the estimated width of the wall from this modality varies slightly across experiments (about $0.04$ m). For the cosine similarity the scores are very high, showing no significant differences across experiments. Additionally, the KDE can be used to compare the distribution of points along the length of the wall, shown in Fig.~\ref{fig:kde_wall1_round1} and Fig.~\ref{fig:kde_wall1_full} for Experiments 1 and 4, respectively. From this, it is evident that the SONAR points do not align perfectly with those of the LiDAR, revealing a shift between the distributions. This indicates a misalignment in the $y$-direction. Finally, the Hellinger distance is calculated, using the KDE density distributions with 100 bins for evaluation across experiments, showing quite high values, indicating a larger difference between the distributions of the LiDAR and the SONARs. In \Cref{fig:interpolation_wall}, the LiDAR and SONAR point clouds are shown overlaid, with the KDE density distributions of both point-clouds depicted beneath.

\section{Conclusion}
The paper presented a unified, GNSS-independent pipeline that fuses LiDAR–IMU with a dual orthogonal forward-looking sonar suite to produce consistent seabed-to-sky maps from an ASV. On the acoustic side, we (i) extended orthogonal wide-aperture stereo fusion to arbitrary rigid translations between the two FLS units -- enabling heterogeneous, non co-located sonars -- and (ii) extracted a leading-edge in each sonar image to generate line-scan constraints. On the mapping side, we modified LIO-SAM to ingest both stereo-derived 3D sonar points and leading-edge line-scans at and between keyframes via motion-interpolated poses, allowing asynchronous, sparse acoustic updates to contribute to a single factor-graph–based map. 

Results highlight both strengths and limitations. Above water, the LiDAR map aligned well with RTK-referenced features after rigid registration; below water, the fused dual SONAR produced structures consistent with canal walls, but analysis against the LiDAR map revealed a systematic wall-width underestimation and a lateral offset, reflected in notable Hellinger distances. We attribute these discrepancies to residual extrinsic errors, elevation-angle uncertainty outside the stereo overlap, misalignments in the mounting of the dual SONAR setup, lack of co-calibration of the SONARs, and the present use of sonar constraints outside the optimizer.

Overall, the proposed pipeline demonstrates that a dual SONAR and LiDAR suite, fused through a LIO-style back-end and fed with both stereo points and leading-edge lines, can yield robust seabed-to-sky maps in GNSS-denied coastal settings, closing key gaps between prior orthogonal FLS reconstruction and multi-domain ASV mapping.

\section*{Acknowledgment}
We thank R. Garmund and P. Mariani (DTU Aqua) for experimental support, A. Libonati Brock (Copenhagen Municipality) for harbour access, and DTU ASTA (DTUASTA20) for calibration. GitHub CoPilot was used for code generation, and ChatGPT for text improvements.

\bibliographystyle{IEEEtran}
\bibliography{references}
\end{document}